\documentclass{article} 
\usepackage{iclr2024_conference,times}

\usepackage{amsmath,amsfonts,bm}

\def\eqref#1{equation~\ref{#1}}

\def\1{\bm{1}}

\DeclareMathAlphabet{\mathsfit}{\encodingdefault}{\sfdefault}{m}{sl}
\SetMathAlphabet{\mathsfit}{bold}{\encodingdefault}{\sfdefault}{bx}{n}

\usepackage{hyperref}
\usepackage{url}
\usepackage{graphicx}

\title{Mentality: A Mamba-based Approach towards Foundation Models for EEG}

\author{Saarang Panchavati, Corey Arnold \& William Speier \\
Department of Radiology\\
University of California Los Angeles\\
Los Angeles, CA 90024, USA \\
\texttt{\{saarang,cwarnold,speier\}@g.ucla.edu} \\
}

\iclrfinalcopy 
\begin{document}

\maketitle
\begin{abstract}
This work explores the potential of foundation models, specifically a Mamba-based selective state space model, for enhancing EEG analysis in neurological disorder diagnosis. EEG, crucial for diagnosing conditions like epilepsy, presents significant challenges due to its noisy, high-dimensional, and nonlinear nature. Traditional machine learning methods have made advances in automating EEG analysis but often fail to capture its complex spatio-temporal dynamics. Recent advances in deep learning, particularly in sequence modeling, offer new avenues for creating more generalized and expressive models capable of handling such complexities. By training a Mamba-based model on a large dataset containing seizure and non-seizure EEG recordings through a self-supervised reconstruction task followed by a seizure detection task, we demonstrate the model's potential, achieving an AUROC of 0.72 on a held-out test set. This approach marks a significant step toward developing large-scale, clinically applicable foundation models for EEG data analysis.

\end{abstract}
\section{Introduction}

Electroencephalography (EEG), a vital, noninvasive tool for recording brain electrical activity, plays a key role in diagnosing and treating neurological disorders such as epilepsy \citep{ACHARYA2013147}. Scalp EEG signals are noisy measurements of brain activity and are characterized by high inter-patient variability, making it challenging to analyze and interpret them \citep{ACHARYA2013147}. Current clinical practice requires time-consuming, manual inspection for diagnosis \citep{ACHARYA2013147}. 

While traditional machine learning models have helped to automate EEG analysis \citep{8972542}, they typically struggle with the high-dimensional, non-linear nature of the signal and fail to capture its spatio-temporal dynamics effectively. More expressive deep learning-based models may offer a solution to improve EEG analysis. In particular, foundation models are large-scale deep learning models trained on extensive datasets to generalize across various tasks. Such models have been proposed in medicine \citep{moor2023foundation}, but work on developing these large models in a clinical context remains limited.

Recent advances in sequence modeling, specifically with selective state space modeling offer a unique opportunity for novel approaches to foundation models applied to biomedical data such as EEG. In particular, Mamba \citep{gu2023mamba} has emerged as a powerful selective state space model with comparable performance to transformer-based approaches on language modeling, genomics, and audio.

In this work, we aim to build a foundation model for EEG, first by relaxing the problem to a more focused context. We train a Mamba-based model on a large public dataset that contains both seizure and non-seizure EEG recordings, first on a self-supervised reconstruction task, and then a downstream seizure detection task.








\section{Methods}






\subsection{Data}

We train and test our data on the Temple University Hospital EEG Seizure Corpus (TUSZ) v2.0.1 \citep{tuhsz}, which is one of the largest annotated seizure corpora to date. The training data subset contains 579 patients, with 2,138 seizure events. The testing subset contains a disjoint set of 43 patients, with 469 seizure events. We included 19 channels from the standard 10-20 EEG montage, all referenced to a reference electrode. All signals are resampled to 200 Hz, and we remove 60 and 120 Hz interference via a notch filter. Data are segmented into 10s non-overlapping windows. Windows are classified as “seizure” or “not seizure” if there is a seizure event in a window.

\begin{figure}[h]
\begin{center}
\includegraphics[scale = .5]{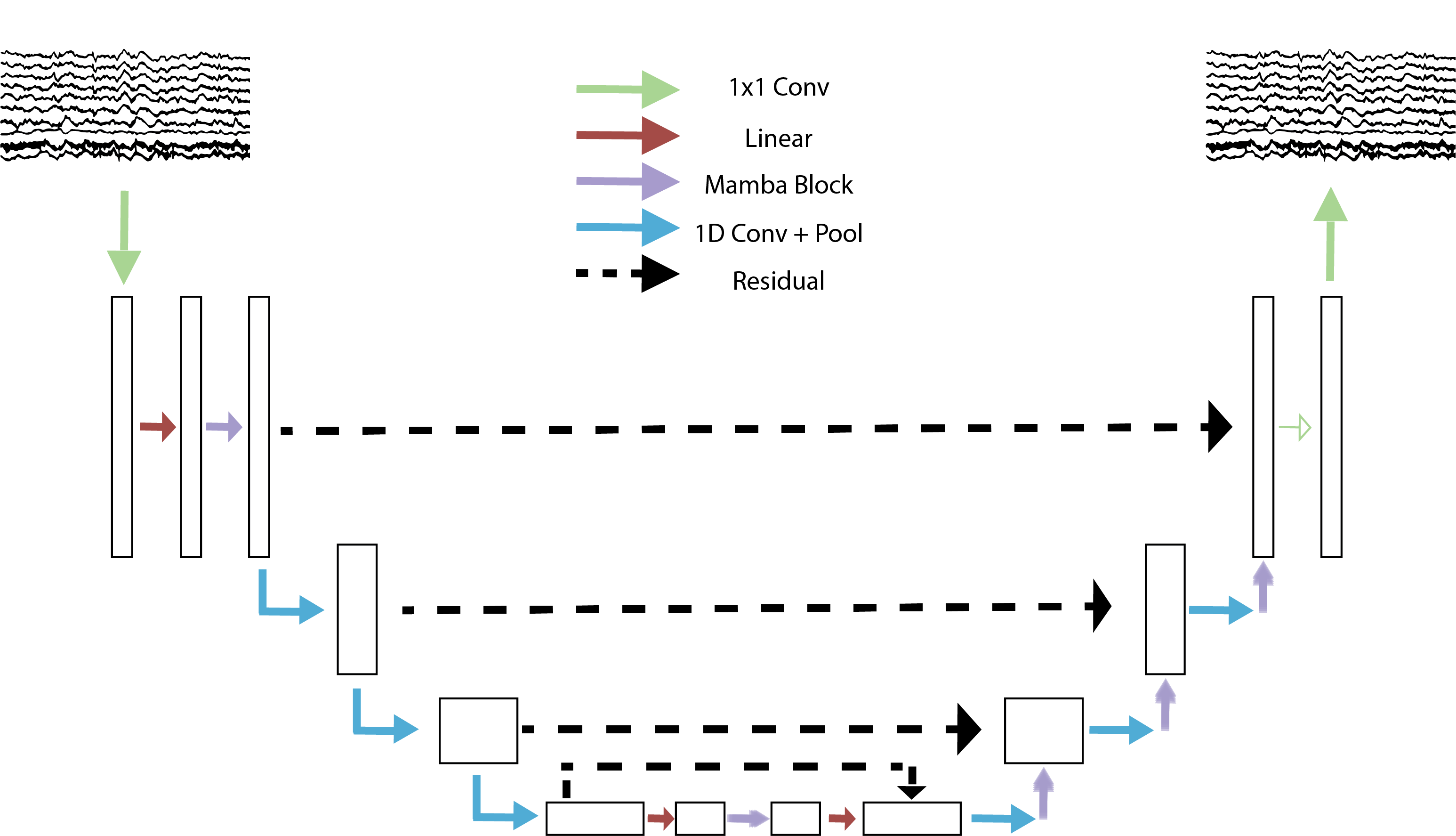}
\end{center}
\caption{Proposed Model Architecture. Inspired by SaShiMi, U-Net, and EEGNet, the model learns a state space representation of frequency-based filters, before downsampling to a smaller representation. The model upsamples back to reconstruct the signal, learning state space representations at each level of sampling till a final output.}
\end{figure}

\subsection{Model Architecture and Training}

Mamba is a deep selective state space model proposed by \citet{gu2023mamba}, capable of analyzing long-range dependencies in sequences. Mamba has been shown to outperform many state-of-the-art tasks in long-context modeling including audio and language. In this work we utilize Mamba “blocks”; which include a Mamba layer followed by layer normalization and a residual connection, which can be stacked together similarly to transformer blocks. In this work, we take inspiration from EEGNet \citep{lawhern2018eegnet}, and SaShiMi \citep{goel2022its} for a full encoder/decoder architecture with Mamba blocks. EEGNet is a lightweight CNN-based model for processing EEG data. SaShiMi is a U-Net \citep{ronneberger2015u} like architecture interspersed with S4 blocks (a precursor to Mamba), where downsampling and upsampling are done by reshaping rather than repeated convolution and pooling. 

Our model begins with a 1D CNN layer of kernel size 100 (half the sampling rate of the signal) to learn “frequency-based” filters up to 50 Hz from each channel. Subsequently, a “channel mixing” linear layer is applied to try to learn the relationships between the channels. This is followed by several Mamba blocks to learn a state space representation of the data, emphasizing the temporal dynamics. 

The output is then downsampled through the traditional U-Net block, composed of a double convolution and mean pooling. We then pass the downsampled input through another set of Mamba blocks, to obtain a downsampled hidden representation of the data. During reconstruction, this is then upsampled through repeated Mamba-transpose convolution - -double convolution blocks. Following the strategy of the U-Net, we concatenate the output of the corresponding downsampling layer to form residual connections with each upsampling layer output to preserve features. We also add a final convolutional layer initialized with zeros to improve sparsity in the final output.

In the downstream task, the downsampled hidden representation is max-pooled along the temporal dimension and then passed through linear layers to obtain a classification probability.

To improve the reconstruction, we employ a combination of mean-squared-error (MSE) loss with a spectral loss, which computes the loss in the Fourier domain. 


\section{Results}
\subsection{Pretraining}

Our best model achieves a MSE of 0.0063. Removing the spectral loss resulted in a four-fold increase in MSE (0.025). Figure 2 demonstrates an example of the reconstruction on an example in the test data.

\begin{figure}[h]
\begin{center}
\includegraphics[scale = .6]{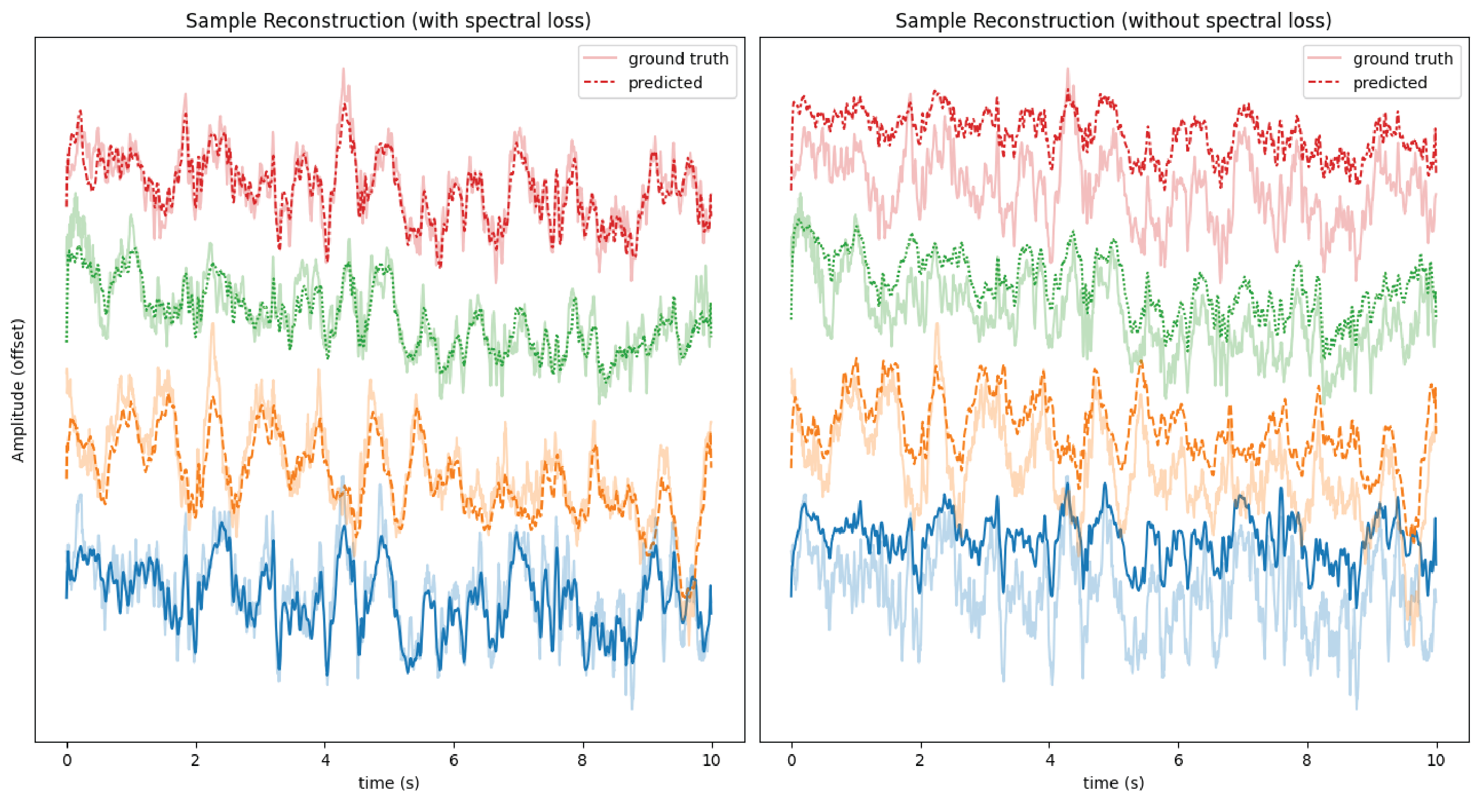}
\end{center}
\caption{Example reconstruction of input sequence on a test sample. Predicted time series are dashed, while ground truth are solid lines. The model captures the general patterns in the time series very well but sometimes fails to completely capture higher frequency components of the signals. The model with spectral loss reconstructs the signal much better than that without.
}
\end{figure}

\subsection{Seizure Detection}

We find that using a pretrained model followed by two linear layers, we can achieve a detection AUROC of 0.72. When we use a model trained from scratch, the detection AUROC drops to 0.64. 

\subsection{Model Interpretability}
A limitation of Mamba blocks is that they do not have a way to analyze the interpretability of their representations. In this work, we rely on analyzing model weights and saliency maps for interpretability. To identify channel importance we compute channel-wise saliency based on particular inputs. Figure 3 demonstrates a manner of highlighting channel importance for a seizure sample. We can also look at the weights of the initial convolutional layer (similar to the EEGNet approach) to identify what types of patterns or frequencies are activated based on the input. 

\begin{figure}[h]
\begin{center}
\includegraphics[scale = .5]{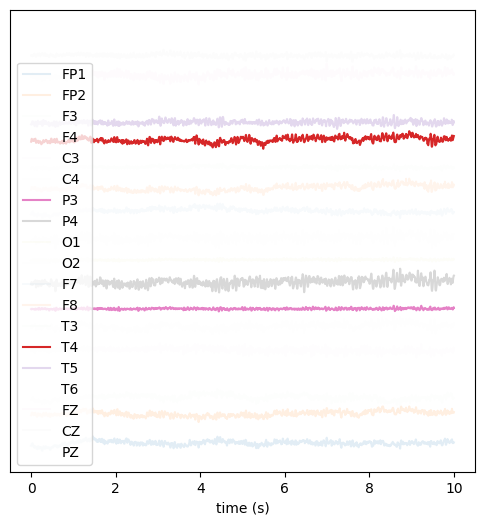}
\end{center}
\caption{Sample seizure window with channel-wise saliency. Channel opacity is a function of its importance to the output. Here, channels T4 and P4 are highlighted as being particularly important to classify the sample as a seizure. We note that these channels exhibit the high-frequency signals characteristic of many seizures. 
}
\end{figure}






\section{Discussion}

This work presents a first step towards developing foundation models for EEG and neural data. Our preliminary results demonstrate the promise that Mamba-based models have for analyzing neural data.  We find that the inclusion of a spectral loss significantly improves the performance of the pretrained model. We also find that pretraining boosts the downstream AUROC significantly, pointing to the importance of this self-supervised pretraining approach. 

One continued drawback of most deep learning approaches to EEG is that while the temporal dynamics of the signal are captured well, the spatial relationships between channels are not meaningfully incorporated into the model. State space models like Mamba process each channel independently. In the approach presented here, we learn a “mixing” of the channels, but this does not necessarily capture the more rigid geometric and functional connections between channels. For example, future works will incorporate a graph-based approach to learning spatio-temporal dynamics at different time windows, similar to \citet{pmlr-v209-tang23a}. Graphs can be constructed both functionally (based on connectivity metrics such as coherence or Granger causality) and/or geometrically, based on spatial distance between electrodes. Exploring these representations will enable the model to extend to various EEG setups and geometries.

Another significant challenge in EEG analysis is the variability in channel configurations across subjects and datasets. Many previous studies address this issue by standardizing the number of channels or interpolating missing channels. These approaches are not geared towards using data from different applications, hardware, and institutions, which contain greater variability and valuable within-patient patterns. In lieu of this, we propose a masked training approach where channels are randomly excluded during model training. This approach will force the model to learn a robust representation based on the available channels. This approach to training along with a structured graph-based architecture can help enhance this model’s usefulness across different setups and in real-world settings such as wearable EEGs.



\subsection{Future Directions}
To expand this model’s applicability, we plan to extend the pretraining stage to a variety of neurological conditions beyond seizure detection. This will include datasets from larger EEG corpora such as the larger TUH EEG Corpus \citep{obeid2016temple} and the UNM Predict+CT \citep{pred_ct}, which cover a range of disorders like Parkinson’s Disease, along with healthy control subjects. This will enhance the range of downstream tasks that this foundation model is capable of, as well as improve the model's ability to learn neural signatures characteristic of these conditions. 

In this approach, we do not explicitly use the learned state-space dynamics in our model (we just use the final hidden representation output of the Mamba blocks). Incorporating these dynamics more explicitly may offer further insights into the temporal dynamics of the EEG signals. These dynamics, combined with the graph approach could greatly improve the interpretability of this model. Clinicians may be able to better understand how channel interactions at a particular time point may contribute to a certain neurological phenomenon. 

As an extension of detection and prediction, this approach may also be extended to help learn relationships between behavioral and neural data, similar to CEBRA \citep{schneider2023learnable}, which learns a joint latent representation between neural data and behavioral data. Mamba’s ability to learn state space representations between sequences offers a similar opportunity to learn related dynamics, perhaps in an even more expressive manner.

\subsection{Conclusions}
The development of a foundation model for EEG data has several practical implications. Such a model would enable researchers and clinicians to more robustly elucidate information from noisy EEG data by adapting to various EEG configurations and formats. This work is an initial step towards that end. Expanding the training data to better cover a range of neurological conditions along with more advanced models will further enhance this ability to analyze neural activity. Such a model could make tangible differences in diagnosis and developing treatments, making an impactful improvement to the understanding and care of neurological disorders.

\subsubsection*{Acknowledgments}
The authors would like to acknowledge Dhruva Karkada and Ekaterina Redekop for their valuable insights in developing this work.

\bibliography{iclr2024_conference}
\bibliographystyle{iclr2024_conference}

\end{document}